# PERFORMANCE ANALYSIS OF WEIGHTED LOW RANK MODEL WITH SPARSE IMAGE HISTOGRAMS FOR FACE RECOGNITION UNDER LOWLEVEL ILLUMINATION AND OCCLUSION

*K.V.Sridhar and Raghu vamshi Hemadri*
*Department of Electronics and Communication Engineering*
*NIT Warangal*

*Abstract* – In a broad range of computer vision applications, the purpose of Low-rank matrix approximation (LRMA) models is to recover the underlying low-rank matrix from its degraded observation. The latest LRMA methods - Robust Principal Component Analysis (RPCA) resort to using the nuclear norm minimization (NNM) as a convex relaxation of the non-convex rank minimization. However, NNM tends to over-shrink the rank components and treats the different rank components equally, limiting its flexibility in practical applications. We use a more flexible model, namely the Weighted Schatten p-Norm Minimization (WSNM), to generalize the NNM to the Schatten p-norm minimization with weights assigned to different singular values. The proposed WSNM not only gives a better approximation to the original low-rank assumption but also considers the importance of different rank components. In this paper, a comparison of the low-rank recovery performance of two LRMA algorithms- RPCA and WSNM is brought out on occluded human facial images. The analysis is performed on facial images from the Yale database and over own database , where different facial expressions, spectacles, varying illumination account for the facial occlusions. The paper also discusses the prominent trends observed from the experimental results performed through the application of these algorithms. As low-rank images sometimes might fail to capture the details of a face adequately, we further propose a novel method to use the image-histogram of the sparse images thus obtained to identify the individual in any given image.  Extensive experimental results show, both qualitatively and quantitatively, that WSNM surpasses RPCA in its performance more effectively by removing facial occlusions, thus giving recovered low-rank images of higher PSNR and SSIM.

## I. INTRODUCTION

The application of low-rank matrix approximation algorithms has proved to give satisfactory and acceptable results in the field of denoising of images. Methods based on Singular value decomposition (SVD) have been formulated and used in image denoising. In [7], a SVD-based method was proposed using non-local, self-similarity, and low-rank approximation, which was tested by adding noises at different levels to standard images. K-means SVD, another advancement of SVD, shows better performance in denoising of images where noise is dominant (low SNR). [8] gives a detailed study of K-SVD for image denoising. Principal component analysis (PCA) is applied for facial color image denoising using the detection of noisy pixels [9]. Robust Principal component analysis (RPCA) was proposed by Candes in [10] and implements the decomposition of a data matrix (for example, an image) into low-rank and sparse matrices. Keeping in mind the complex nature of noise, a more generative version of RPCA was proposed in [12] modelling noise as a mixture of Gaussians(MoG). [1] gives a broad overview of the application of RPCA in the field of image denoising and video background separation. The application of fixed-rank RPCA in the field of background recovery is explored in [11].

The recent increase in the importance of facial recognition for almost all security purposes subsequently directs many researchers in facial image denoising and facial modeling field using the same LRMA concepts. Proper identification of the person in a given image is possible when the image is occlusion-free. In the presence of noise or occlusions, the recognition algorithms fail to identify the person in the images. Thus, pre-processing of the image to produce a more precise image becomes necessary before recognition. Application of RPCA in color facial images denoising and the comparison of its performance with that of other state-of-the-art denoising algorithms is done in [2]. An approach to learn a structured low-rank representation for image classification is presented in [19].  The occlusion free low-rank image can then be used for facial recognition, but many important details are lost making them a bit blurry. So instead, usage of sparse components and their intensity for identification of the person was proposed in [3]. Face recognition from Sparse representation was executed by J. Wright in [6]. [5] proposes a method by assuming noise as a Mixture of Exponential power (MoEP), which can deal with complex noises other than Gaussian and Laplacian noises, giving good results in facial modelling. Further research led to the proposal of another advancement of RPCA- weighted Schatten p-norm (WSNM). It was observed in [4] that in comparison to state-of-the-art methods, WSNM removes noise more effectively and can model dynamic and complex situations better. Another variant of WSNM using ADMM method was proposed in [16]. WSNM is basically a generalised version of Weighted Nuclear Norm Minimisation whose performance in image denoising was analysed in [13],[14].

## II. DESCRIPTION OF ALGORITHMS

### A. RPCA

Robust Principal Component Analysis (RPCA) is a modification of the statistical procedure PCA and works well under grossly corrupted or noisy observations. RPCA aims to procure a low-rank version and a sparse version from a single matrix and finds extensive applications in image denoising, image modelling, background separation in images, videos, and several machine learning purposes. There are many ways of implementing RPCA – ADMM, exact and inexact Augmented Lagrange's Multipier(ALM). A data matrix M can be decomposed as:

$$M=L+S$$

Where L is the low-rank matrix and, S is the sparse matrix. L can be further decomposed as:

$$L=U\sum V'$$

Singular value decomposition, where $\sum$ contains the singular values along its diagonals in decreasing order. The low-rank matrix is found out using its nuclear norm:

$$//L//_* = ||\sum||_1$$

Usually, the cardinality of S is supposed to be minimum for the optimum solution of L and S; hence initially, it was proposed that the optimum S will be found through its of l0 norm. However, the l0 norm is non-convex and computationally NP-hard. Thus, the sparse matrix is found out using the l1 norm of S as the l1 norm is the tightest convex approximation of l0 norm. Hence, RPCA finds:

$$\min(//L//_* + \lambda //S//_1), \text{ s.t. } M=L+S$$

Thus L*, S* are calculated (the approximated values of L and S) such that:

$$L^*, S^* = \underset{L, S}{\arg\min} //L//_* + \lambda //S//_1, \text{ s.t. } M-L-S=0$$

Inexact ALM is preferred over exact ALM due to its fast convergence rate. RPCA method is robust against outliers but is currently computationally expensive. Moreover, the method assigns equal weights to all the singular values or rank components. Thus, its flexibility in practical applications is limited. WSNM, on the other hand, optimally assigns different weights to different rank components leading to more flexibility and better results in practical applications like facial image denoising.

*B. WSNM*

Low-rank approximation tends to use nuclear norm minimization (NNM) as the nuclear norm is the tightest convex relaxation of rank minimization issue. Given a matrix Y, NNM finds a low-rank matrix X that follows the equation:

$$X = \underset{x}{\arg\min} \| X - Y \|_F^2 + \lambda ||X||_*$$

Where $\lambda$ is a trade-off parameter between the loss function of the nuclear norm induced LRMA. Unlike NNM, Weighted Schatten p-norm minimization assigns different weights to different rank components. Image denoising, background subtraction that generally evolves around LRMA, has a great application of WSNM. Weighted nuclear norm is given as:

$$\| X \|_{\omega, *} = \sum_i |\omega_i \sigma_i(X)|_1$$

Where $\omega = [\omega_1, \omega_2, ..., \omega_n]$ and $\omega_i \geq 0$ and assigned to $\sigma_i(X)$. The algorithm of low-rank and sparse decomposition in WSNM has been described as follows:

In NNM, main aim is to decompose a given matrix Y into a low-rank(X) and an error matrix (E): Y=X+E following:

$$\underset{E, X}{\min} \| E \|_1 + ||X||_*$$

Whereas in WSNM, we use the loss function:

$$\underset{E, X}{\min} \| E \|_1 + \| X \|^P_{\omega, SP} \text{ such that } Y=X+E.$$

Using Augmented Lagrangian function, we get:

$$L(X, E, Z, \mu) = \|E\|_1 + \|X\|^P_{\omega,SP} + \langle Z, Y-X-E \rangle + \mu/2 \|Y-X-E\|_F^2$$

where Z is Lagrangian multiplier, μ is positive scalar. The weighted vector values are defined as
$$\omega_i = C\sqrt{(m \times n)}/(\sigma_i(Y)+\epsilon)$$

## III. EXPERIMENTS AND VISUAL RESULTS

Facial images of first 6 individuals (11 images of each person) from Yale database depicting pictures under shadows, illuminations, facial expressions were used for the experiments. The images of a person were first cropped to focus only on the facial part such that they had the same dimensions. The pictures belonging to the same person were stacked into a single matrix as its columns, the pixel values of an image being the contents of each column. These matrices were subjected to RPCA and WSNM algorithms for recovering low-rank and sparse versions of 11 images at once. Occlusions (expressions, shadows, etc.) are usually sparse, which get separated in our sparse matrices, whereas the necessary facial features being correlated and redundant, form the low-rank matrix.

The data matrix obtained after stacking the images into its columns is of a higher rank, and the lower rank versions are obtained by applying different values of the hyper-parameters to get different ranks. To interpret which rank gives us the best results, a quantitative analysis of the results (of different ranks for L matrix) was performed. The values of Peak Signal to Noise Ratio (PSNR), Structural Similarity Index (SSIM) were observed for some of the images under different ranks, and then the best results were identified. PSNR defining how much the image is free of noise corruption is expected to have a high value for good results. SSIM being the similarity index always lies between values 0 and 1; the closer the value of SSIM to 1, the closer is the result of the original data. For RPCA, the values of hyper-parameters-lambda, mu, error threshold were tuned to optimum values. For WSNM, we considered p-norm where p=0.95 and the parameters- lambda, mu, error tolerance, patch-size (taken as a constant), noise-variance (taken as constant) were tuned to have optimum values. The recovered low-rank and sparse versions of four images of subject one from Yale database after application of RPCA is presented in Fig. 1. No matter what the occluded image be, the lower-rank version approximately produces the same basic facial image for each case, as can be seen in Fig.1, whereas the occlusions get separated in the sparse images.

## IV. OBSERVATIONS

The PSNR of the low-rank versions of different subjects for any occlusion were seen to increase to the highest PSNR value and then again reducing with the increase of rank.

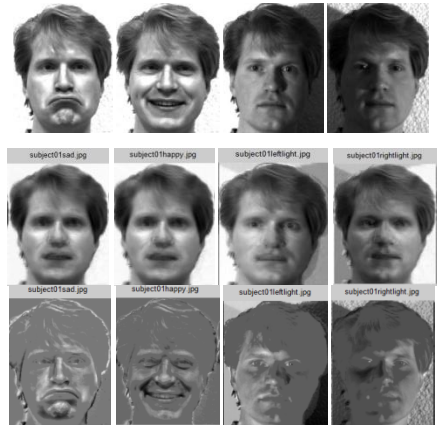

**Fig. 1.** The original images, low-rank images and sparse versions (from top to bottom) of person 1 (from Yale database) being sad, happy, and with shadows on face due to improper location of light source (from left to right)

At extremely low ranks usually the images lose clarity of distinct facial features and become blurry, leading to poor PSNR, though they are devoid of occlusions. Again as rank increases apart from basic essential features (enough to develop a proper facial image), some unwanted occlusions or outliers also get added to low-rank images, so images lose their authenticity and show low PSNR at high ranks. It was observed both in the case of RPCA and WSNM results. Fig. 2 depicts this trend using RPCA results.

PSNR and SSIM, the two parameters are inter-dependent. As can be seen in Fig 3, the SSIM curve follows the PSNR curve's shape. When the low-rank image is structurally similar to the original occlusion-free image, the noise content or occlusion in the restored low-rank image is less.

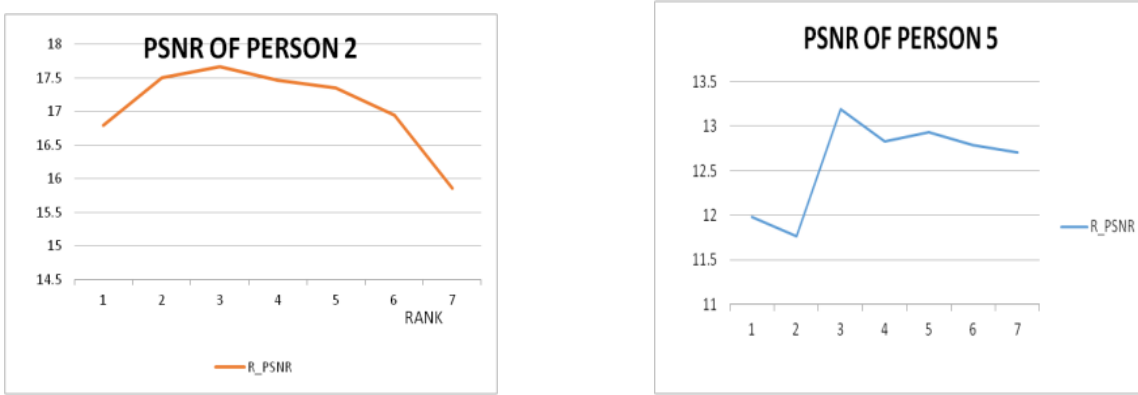

**Fig. 2.** PSNR curves of person 2 and 5 obtained by applying RPCA to their respective images with shadow occlusions due to left-side light source.

RPCA treats all singular values equally unlike WSNM. WSNM is undoubtedly expected to give distinct images with more facial features intact in the low-rank components. Highest WSNM –PSNR values were always at least 2 to 3 dB better than the highest RPCA-PSNR values (Fig. 3). SSIM values were also better in case of WSNM than RPCA.

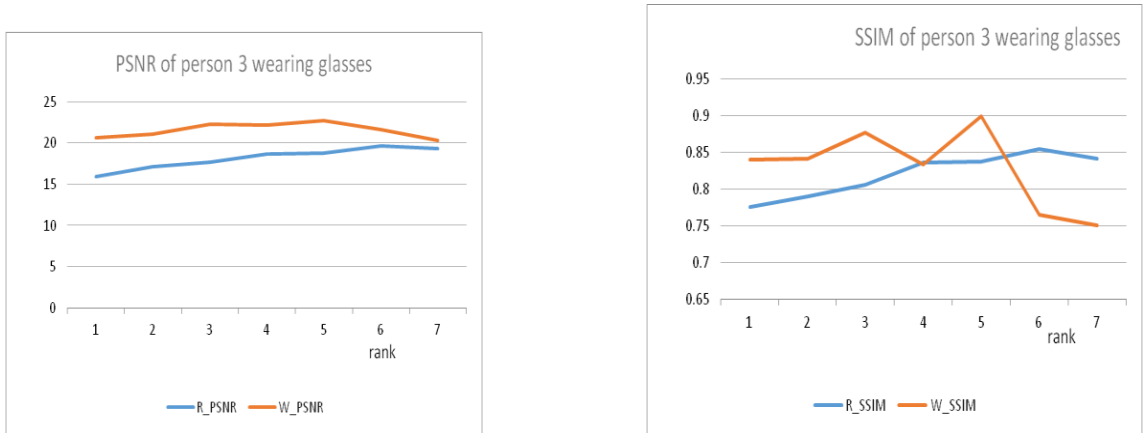

**Fig. 3** PSNR and SSIM curves of person 3 with glasses as occlusion with highest PSNR with RPCA being 19.672 db and highest PSNR with WSNM being 22.738 clearing showing the enhancement in WSNM performance. Highest RPCA-SSIM value was 0.854 which gets enhanced to 0.899 with WSNM.

The ranks with best results after applying RPCA and WSNM are listed in Table 1 for six persons under some of the occlusions. The rank with the best results is not same in all cases, and totally depends on the inherent facial features of a person, the type of occlusion present in the facial images. For RPCA, there was no proper visible trend of similarity either across different persons for the same occlusion or for the same person under different occlusions. However, in WSNM, for each occlusion, one of the ranks was seen to have the best results for numerous persons. For example, for left-light occlusion, rank one was seen to give the best results for 5 out of 6 persons (except person 2). For sad expression, rank 3 gives best results for 4 out of 6 persons (except persons 1 and 2). Similar kinds of occlusions showed similar kinds of PSNR and SSIM curves irrespective of persons. Similarity was observed between curves of images with happy, sad, sleepy, surprised, wink expressions (facial expressions).

**Table 1**. Ranks with best results in experiments on Yale database

| Wearing glasses | P1 | P2 | P3 | P4 | P5 | P6 |
|---|---|---|---|---|---|---|
| Rank with best results-RPCA | 3 | 4 | 6 | 3 | 8 | 5 |
| Rank with best results-WSNM | 4 | 1 | 5 | 4 | 4 | 6 |
| Happy expression | P1 | P2 | P3 | P4 | P5 | P6 |
| Rank with best results-RPCA | 3 | 4 | 6 | 5 | 4 | 8 |
| Rank with best results-WSNM | 1 | 5 | 3 | 3 | 3 | 8 |
| Sad expression | P1 | P2 | P3 | P4 | P5 | P6 |
| Rank with best results-RPCA | 3 | 4 | 6 | 5 | 3 | 8 |
| Rank with best results-WSNM | 4 | 1 | 3 | 3 | 3 | 3 |
| Shadow occlusion due to left-light | P1 | P2 | P3 | P4 | P5 | P6 |
| Rank with best results-RPCA | 2 | 3 | 6 | 2 | 3 | 2 |
| Rank with best results-WSNM | 1 | 2 | 1 | 1 | 1 | 1 |
| Shadow occlusion due to right-light | P1 | P2 | P3 | P4 | P5 | P6 |
| Rank with best results-RPCA | 2 | 3 | 1 | 5 | 4 | 3 |
| Rank with best results-WSNM | 3 | 3 | 2 | 2 | 2 | 2 |

P1, P2, P3, P4, P5, P6- person 1, 2, 3, 4, 5, 6.

## V. EXPERIMENTS ON CREATED DATABASE

We created a database containing pictures of 2 subjects: one of our authors under different illumination levels, and another with occlusions like different expressions, glasses- Fig.5(a),(b). Applying WSNM and RPCA on the created database, the occlusion-free lower rank versions were recorded along with the sparse images together with PSNR, SSIM. As expected, WSNM performance was seen to be better than that of RPCA (at least by 1 or 2 dB in PSNR). PSNR, SSIM values under different ranks were always better in the case of WSNM, Fig. 4 presents the plots in support of this observation. Similar trends were observed in PSNR and SSIM curves (refer Fig. 4).

**Table 2.** Ranks with best results for experiments on created database

| ILLUMINATION OCCLUSIONS | | |
|---|---|---|
| Cases | Rank with best results-RPCA | Rank with best results-WSNM |
| frontlight 1 source | 2 | 2 |
| frontlight 2 sources | 3 | 2 |
| leftlight 1 source | 2 | 2 |
| leftlight 2 sources | 3 | 2 |
| toplight 1 source | 4 | 3 |
| toplight 2 sources | 4 | 2 |
| rightlight 1 source | 2 | 2 |
| rightlight 2 sources | 2 | 2 |
| GLASSES AND EXPRESSIONS AS OCCLUSIONS | | |
| Cases | Rank with best results-RPCA | Rank with best results-WSNM |
| happy | 1 | 1 |
| happy with glasses | 4 | 3 |
| sad | 1 | 1 |
| sad with glasses | 3 | 1 |
| sleepy | 3 | 3 |
| sleepy with glasses | 4 | 1 |
| wink | 2 | 1 |
| wink with glasses | 3 | 1 |

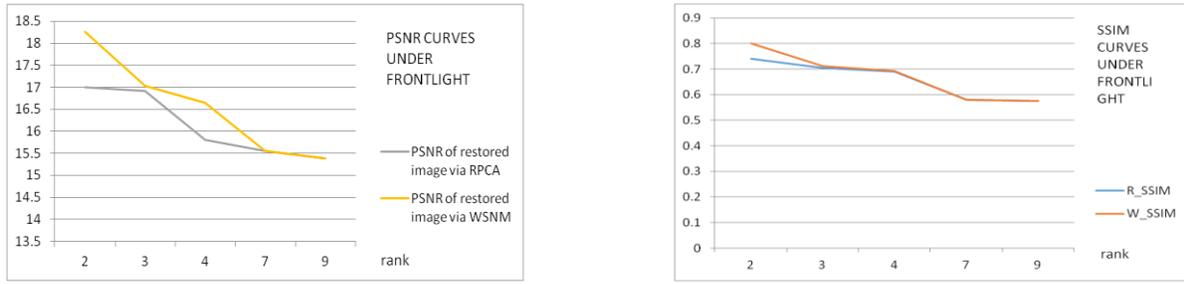

**Fig. 4**   PSNR and SSIM curves of photos of person 1 under front light illumination from one source. The highest PSNR value observed under front-light illumination at rank 1 was 18.254 db for WSNM and 17.001 db for RPCA. The highest SSIM value observed under front-light illumination at rank 1 was 0.8 in case of WSNM whereas it was 0.74 in case of RPCA.

Experiments done on the illumination occlusions, with one and two light sources from the front, left, right, top, exhibited that for RPCA either rank 2, 3, or 4 gave the best results. But, WSNM gave the best results at mainly rank 2, as is seen from Table 2. This is because these images all had the same kind of occlusions- shadows due to varying light-source position. Similarly, for experiments on different facial expressions, WSNM produced the best results at rank 1 in maximum cases. showing more consistency in its results across different images with the same type of occlusion in it.

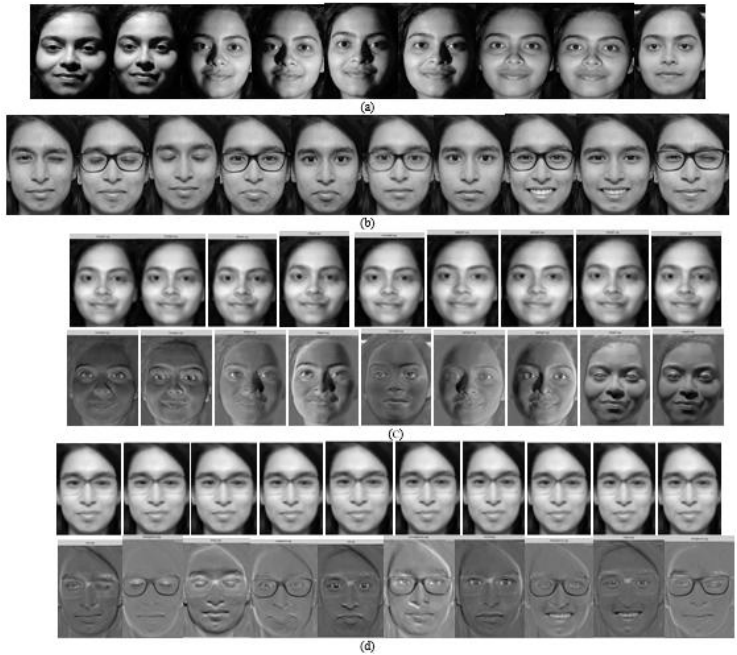

**Fig. 5**(a). Facial Images of Person 1 with light sources from top( 1 and 2 sources), right( 1 and 2 sources), left(1 and 2 sources), front( 1 and 2 sources) and under normal illumination (b) Facial Images of person 2 with happy, sad, sleepy and wink expressions both with and without wearing glasses (c) Low rank(**rank 2**) and sparse versions of the Facial Images of person 1 after applying WSNM (d) Low rank(**rank 1**) and sparse versions of the Images of person 2 after applying WSNM.

V. COMPARATIVE ANALYSIS OF RPCA, WNNM, WSNM

For better insight into the performance comparison of weighted low-rank models, PSNR, SSIM values were recorded for RPCA, WNNM and WSNM (3 values of p) on images from Yale and our database. Optimum results for WSNM were seen at p=0.8. PSNR, SSIM were also recorded at p=0.9 and p=0.95 for WSNM. For most cases, PSNR and SSIM values were better in case of WSNM compared to RPCA and WNNM. From the PSNR, SSIM values for different values of p, it was evident that p=0.95 gives the worst values compared to both p=0.9 and p=0.8. Moreover, the pictures with illumination-related occlusions shows best results at lower ranks whereas the expression-related occlusions lead to best results at higher ranks. WSNM shows more consistency as it shows best results at the same rank for different p values in maximum of the cases for the same person, but RPCA shows best results at random ranks for different cases even for the same person. Better visualisation of the results can be done from Fig. 6 where in each of the 4 cases the low-rank and sparse images obtained through RPCA, NNM and WSNM have been presented, the difference in clarity of images in accordance to the PSNR values is evident.

TABLE 3. PSNR values for several testcases via RPCA, WNNM, WSNM

| | | | PSNR TABLE | | | | |
|---|---|---|---|---|---|---|---|
| | Case | | RPCA | WNNM | WSNM(p=0.8 optimum) | WSNM(p=0.9) | WSNM(p=0.95) |
| YALE DATABASE | Person 1 | Glasses | 19.61(3) | 21.098(4) | 21.768(4) | 21.365(4) | 21.416(4) |
| | | Happy | 19.153(3) | 19.87(3) | 20.149(3) | 20.015(3) | 19.831(1) |
| | | Sad | 19.594(3) | 20.164(3) | 20.995(4) | 20.82(4) | 20.762(4) |
| | | Left-light | 9.627(2) | 10.51(1) | 10.8(1) | 10.664(1) | 10.103(1) |
| | | Right-light | 8.576(2) | 8.316(1) | 9.32(1) | 8.741(2) | 8.241(3) |
| | Person 2 | Glasses | 23.98(4) | 24.716(1) | 25.289(1) | 24.942(1) | 25.101(1) |
| | | Happy | 23.696(4) | 24.716(1) | 25.387(1) | 24.94(1) | 25.331(5) |
| | | Sad | 24.131(4) | 24.796(1) | 25.397(1) | 24.942(1) | 25.101(1) |
| | | Left-light | 17.66(3) | 22.289(1) | 24.628(1) | 24.123(1) | 21.014(1) |
| | | Right-light | 17.695(3) | 22.493(1) | 24.635(1) | 24.186(1) | 19.024(1) |
| | Person 3 | Glasses | 19.672(6) | 23.132(5) | 23.727(4) | 23.617(4) | 22.738(5) |
| | | Happy | 19.188(6) | 21.72(4) | 21.527(3) | 21.501(3) | 21.393(3) |
| | | Sad | 19.774(5) | 22.76(4) | 23.881(4) | 23.76(4) | 21.608(5) |
| | | Left-light | 9.914(6) | 10.648(1) | 10.959(1) | 10.94(1) | 10.009(1) |
| | | Right-light | 8.375(1) | 9.465(1) | 9.742(1) | 9.628(1) | 9.261(2) |
| | Person 4 | Glasses | 18.999(3) | 21.563(1) | 22.076(3) | 21.768(3) | 21.387(4) |
| | | Happy | 20.844(5) | 23.472(5) | 23.897(5) | 23.753(5) | 21.901(3) |
| | | Sad | 21.04(5) | 24.118(5) | 24.44(5) | 23.443(4) | 22.566(3) |
| | | Left-light | 15.116(2) | 20.123(1) | 21.896(1) | 19.833(1) | 21.256(1) |
| | | Right-light | 12.067(5) | 13.756(2) | 13.872(1) | 13.725(2) | 13.845(2) |
| | Person 5 | Glasses | 23.134(8) | 22.407(1) | 22.775(1) | 22.646(1) | 22.743(4) |
| | | Happy | 21.687(4) | 22.407(1) | 22.879(1) | 22.746(1) | 22.636(3) |
| | | Sad | 21.601(3) | 22.407(1) | 22.81(1) | 22.531(1) | 22.686(3) |
| | | Left-light | 13.195(3) | 16.162(1) | 16.402(1) | 16.215(1) | 15.157(1) |
| | | Right-light | 10.012(4) | 10.842(1) | 13.50(1) | 13.322(1) | 11.153(2) |
| | Person 6 | Glasses | 19.75(4) | 22.996(4) | 23.35(4) | 23.008(3) | 22.966(6) |
| | | Happy | 26.323(8) | 26.084(5) | 28.657(5) | 27.018(5) | 27.98(8) |
| | | Sad | 23.651(8) | 25.897(5) | 28.134(5) | 26.489(5) | 23.33(6) |
| | | Left-light | 15.413(2) | 20.575(1) | 21.028(1) | 19.944(1) | 20.372(1) |
| | | Right-light | 18.55(3) | 20.65(1) | 21.078(1) | 20.664(1) | 20.384(2) |
| OUR DATABASE | Person 1 | frontlight 1 source | 17.001(2) | 18.362(2) | 18.763(2) | 18.254(2) | 18.834(2) |
| | | frontlight 2 souces | 17.88(2) | 18.177(2) | 18.602(2) | 18.062(2) | 18.682(2) |
| | | leftlight 1 source | 14.889(2) | 16.516(1) | 16.705(1) | 16.419(2) | 16.372(1) |
| | | leftlight 2 sources | 14.75(2) | 15.894(1) | 16.011(1) | 15.667(2) | 15.76(1) |
| | | rightlight 1 source | 14.78(2) | 16.367(1) | 16.598(1) | 16.579(2) | 16.226(1) |
| | | rightlight 2 sources | 15.789(2) | 16.36(1) | 16.564(1) | 16.524(2) | 16.221(1) |
| | Person 2 | happy | 19.85(1) | 21.319(1) | 22.945(1) | 21.66(3) | 21.025(1) |
| | | happy with glasses | 20.44(4) | 22.755(3) | 23.012(4) | 22.126(4) | 21.995(4) |
| | | sad | 15.55(1) | 21.042(1) | 22.301(1) | 21.125(1) | 21.169(1) |
| | | sad with glasses | 19.78(1) | 22.755(1) | 22.315(1) | 21.804(1) | 21.26(1) |
| | | sleepy | 21.44(3) | 21.994(3) | 24.201(3) | 23.68(3) | 23.54(3) |

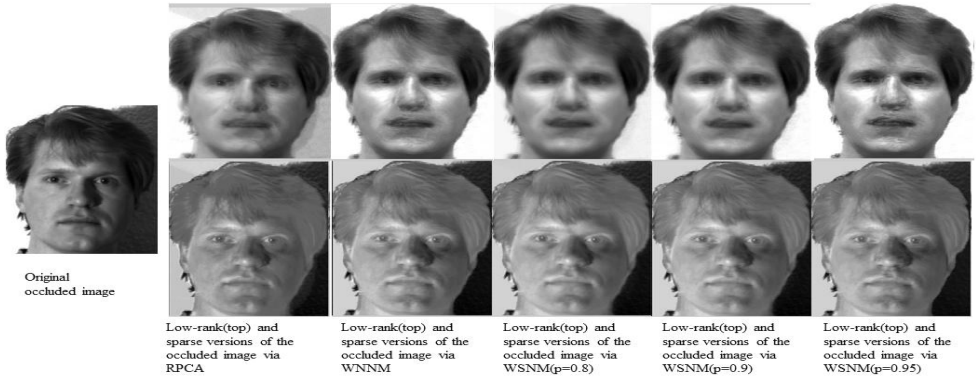

(a)

Original occluded image

Low-rank(top) and sparse versions of the occluded image via RPCA

Low-rank(top) and sparse versions of the occluded image via WNNM

Low-rank(top) and sparse versions of the occluded image via WSNM(p=0.8)

Low-rank(top) and sparse versions of the occluded image via WSNM(p=0.9)

Low-rank(top) and sparse versions of the occluded image via WSNM(p=0.95)

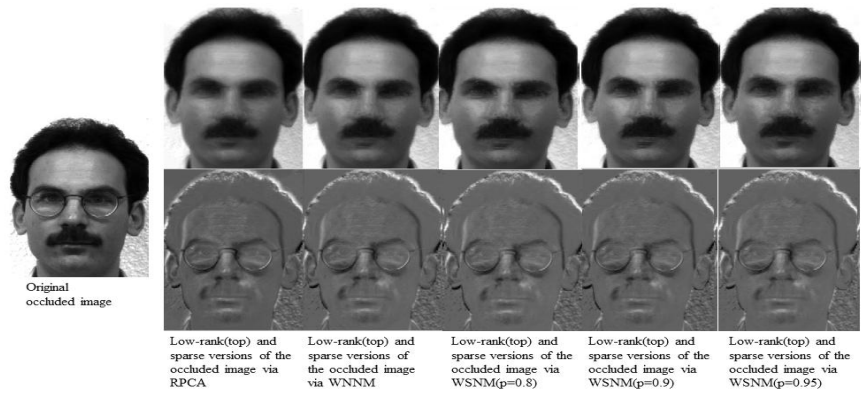

(b)

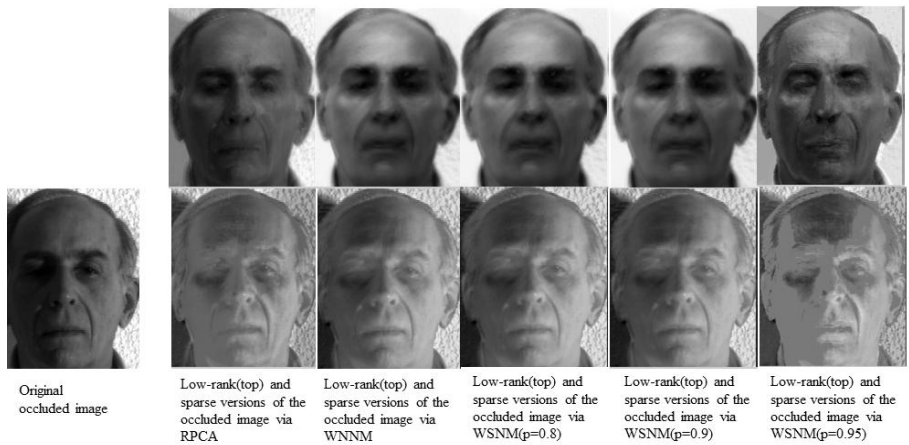

(c)

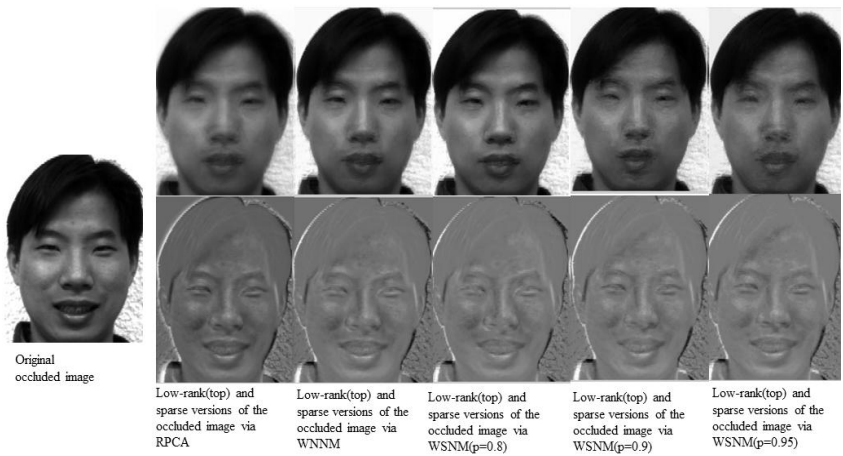

(d)

**Fig. 6** Pictures of the original occluded image, low-rank and sparse images after application of RPCA, WNNM, WSNM(for three values of p -0.8, 0.9, 0.95) for (a) subject 1 left-light case (b) subject 2 glasses case (c) subject 5 right-light case (d) subject 4 happy case from the Yale database

TABLE 4. SSIM values for several cases via RPCA, WNNM, WSNM

| | | | SSIM TABLE | | | | |
|---|---|---|---|---|---|---|---|
| | **Case** | | RPCA | WNNM | WSNM(p=0.8 optimum) | WSNM(p=0.9) | WSNM(p=0.95) |
| YALE DATABASE | Person 1 | Glasses | 0.698(3) | 0.724(4) | 0.749(4) | 0.743(4) | 0.741(4) |
| | | Happy | 0.678(3) | 0.683(3) | 0.69(3) | 0.689(3) | 0.676(1) |
| | | Sad | 0.691(3) | 0.686(3) | 0.732(4) | 0.725(4) | 0.724(4) |
| | | Left-light | .476(2) | 0.56(1) | 0.627(1) | 0.565(1) | 0.545(1) |
| | | Right-light | 0.459(2) | 0.457(1) | 0.482(1) | 0.474(2) | 0.408(3) |
| | Person 2 | Glasses | 0.808(4) | 0.829(1) | 0.834(1) | 0.834(1) | 0.832(1) |
| | | Happy | 0.808(4) | 0.828(1) | 0.834(1) | 0.834(1) | 0.832(5) |
| | | Sad | 0.809(4) | 0.829(1) | 0.85(1) | 0.834(1) | 0.847(1) |
| | | Left-light | 0.759(3) | 0.822(1) | 0.833(1) | 0.832(1) | 0.807(2) |
| | | Right-light | 0.778(3) | 0.822(1) | 0.833(1) | 0.832(1) | 0.788(1) |
| | Person 3 | Glasses | 0.854(6) | 0.903(5) | 0.905(4) | 0.904(4) | 0.899(5) |
| | | Happy | 0.81(5) | 0.83(4) | 0.869(3) | 0.869(3) | 0.867(3) |
| | | Sad | 0.851(6) | 0.893(4) | 0.899(4) | 0.898(4) | 0.877(3) |
| | | Left-light | 0.547(1) | 0.61(1) | 0.626(1) | 0.624(1) | 0.59(1) |
| | | Right-light | 0.447(1) | 0.533(1) | 0.552(1) | 0.546(1) | 0.511(2) |
| | Person 4 | Glasses | 0.746(2) | 0.767(1) | 0.78(3) | 0.777(1) | 0.768(4) |
| | | Happy | 0.788(5) | 0.84(5) | 0.842(5) | 0.8415(5) | 0.788(5) |
| | | Sad | 0.814(5) | 0.854(5) | 0.853(5) | 0.828(4) | 0.807(3) |
| | | Left-light | 0.624(2) | 0.762(1) | 0.789(1) | 0.756(1) | 0.768(1) |
| | | Right-light | 0.567(2) | 0.647(2) | 0.65(1) | 0.559(2) | 0.645(1) |
| | Person 5 | Glasses | 0.77(8) | 0.697(1) | 0.738(1) | 0.705(1) | 0.741(4) |
| | | Happy | 0.68(3) | 0.697(1) | 0.738(1) | 0.705(1) | 0.736(3) |
| | | Sad | 0.68(3) | 0.697(1) | 0.78(1) | 0.76(1) | 0.734(3) |
| | | Left-light | 0.597(3) | 0.668(1) | 0.672(1) | 0.671(1) | 0.657(1) |
| | | Right-light | 0.438(3) | 0.552(1) | 0.672(1) | 0.652(1) | 0.496(2) |
| | Person 6 | Glasses | 0.788(4) | 0.854(4) | 0.859(4) | 0.856(3) | 0.848(3) |
| | | Happy | 0.895(8) | 0.915(5) | 0.932(5) | 0.928(5) | 0.908(8) |
| | | Sad | 0.853(8) | 0.914(5) | 0.917(5) | 0.915(5) | 0.856(3) |
| | | Left-light | 0.706(2) | 0.8(1) | 0.81(1) | 0.801(1) | 0.804(1) |
| | | Right-light | 0.78(3) | 0.801(1) | 0.809(1) | 0.802(1) | 0.804(1) |
| OUR DATABASE | Person 1 | frontlight 1 source | 0.74(2) | 0.812(2) | 0.813(2) | 0.8(2) | 0.815(2) |
| | | frontlight 2 souces | 0.725(2) | 0.812(2) | 0.808(2) | 0.799(2) | 0.814(2) |
| | | leftlight 1 source | 0.679(2) | 0.796(1) | 0.789(1) | 0.792(2) | 0.787(1) |
| | | leftlight 2 sources | 0.748(2) | 0.788(1) | 0.778(1) | 0.757(2) | 0.786(1) |
| | | rightlight 1 source | 0.752(2) | 0.794(1) | 0.796(1) | 0.794(2) | 0.792(1) |
| | | rightlight 2 sources | 0.695(2) | 0.794(1) | 0.796(1) | 0.793(2) | 0.792(1) |
| | Person 2 | happy | 0.741(1) | 0.76(1) | 0.801(1) | 0.786(3) | 0.759(1) |
| | | happy with glasses | 0.741(4) | 0.784(3) | 0.792(4) | 0.79(4) | 0.779(4) |
| | | sad | 0.698(1) | 0.761(1) | 0.791(1) | 0.786(1) | 0.775(1) |
| | | sad with glasses | 0.697(1) | 0.757(1) | 0.791(1) | 0.785(1) | 0.775(1) |
| | | sleepy | 0.815(3) | 0.772(3) | 0.795(3) | 0.782(3) | 0.778(3) |

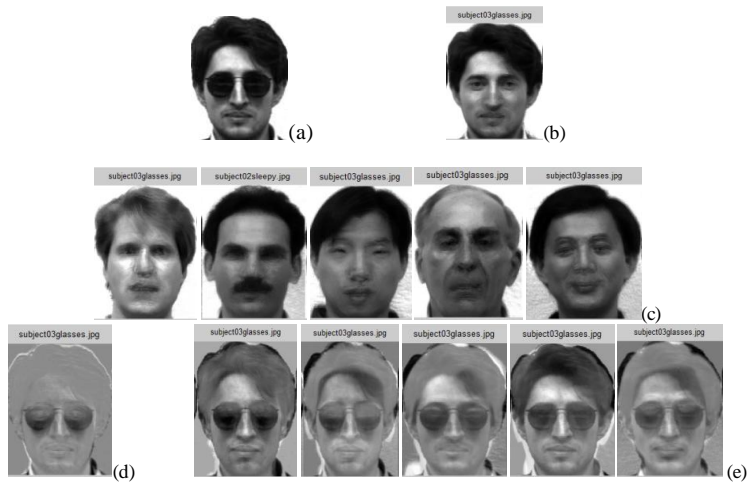

**Fig 7.** (a)Test image of subject 3 wearing glasses from Yale(b) Low-rank obtained when test image subjected to WSNM , stacked with subject 3 images(correct individual) (c) Low-rank images obtained when test image stacked with subject 1 , 2 , 4 , 5 , 6 images(from left to right)and then subjected to WSNM (d) Sparse image obtained when test image subjected to WSNM , stacked with subject 3 images (e) Sparse images obtained when test image stacked with subject 1 , 2 , 4 , 5 , 6 images(from left to right)and then subjected to WSNM

## VII. FACIAL RECOGNITION BY SPARSE IMAGE HISTOGRAM

The algorithms like RPCA, WSNM, SVD, etc. which generate occlusion-free low-rank images along with the sparse versions, are used for pre-processing human facial images before they can be used for face recognition. The low-rank versions being occlusion-free, to some extent, help in better and more accurate identification of the person. Although the low-rank images tend to be free of occlusion, they often turn out to be blurry, which hinders the algorithms from realizing the essential features of the person's face, thus giving wrong results.

During the low-rank recovery of the image, the intricate details of the face are also lost along with the occlusions like edges of the face, sharpness, etc. and those lost details can be vital contributors to proper facial-recognition. Therefore, low-rank images may fail in the proper identification of an individual from the images. Thus, the concept of usage of the sparse images comes into existence in [3]. In [3], the use of intensity of the sparse images (obtained by applying RPCA) has been elaborated through two parameters of sparsity and smoothness, which can be used for identification of the person whose image is being talked about. Here, the histogram of sparse image data is used to determine the individual whose image is being tested. From the Yale database, the photo of subject 3 wearing glasses was chosen as the test image on which WSNM was applied along with 11 images of 6 different individuals (subjects 1, 2, 3, 4, 5, 6) of Yale. Fig 7(a) shows the test image which was stacked with images of 6 persons. In each case, the low-rank and sparse versions derived from the test image after applying with photos of 6 persons were observed (refer Fig. 7). When a given test image that belongs to subject 3 is tested with the images of subject 3 itself, the associated sparse image reflects only the intra-class difference (differences between images of the same person) i.e., shadows and occlusion. When the test image is stacked with images of an incorrect individual, it was observed that due to the low-rank constraint of WSNM, the test image gives faces of the incorrect individuals (Fig 7(c)) in its recovered low-rank versions although it belongs to subject 3. The sparse images contain discriminative person-specific information (Fig 7(e)). Compared to the error image obtained from the correct individual (subject 3) in Fig .7(d), the sparse images from the incorrect individuals in Fig. 7(e), not only contain the intra-class difference (sunglasses) but also contain inter-class information or the differences in images of different individuals. (shadows, outline, ears, chin., hair, etc.).

To obtain much deeper insight into the differences among the sparse error images obtained from the correct and incorrect individuals, the histogram of image data was plotted for every image. A histogram-plot defines equally spaced bins (each representing a range of data values), and then it calculates the number of pixels within each range of values. The data values represent different colors present in the image; the vertical axis represents how many pixels each color contributes to the entire image. Here, the experiments involve grayscale images, and the data values extend from 0 to 1 (where 0 represents black, and 1 represents white). The histograms obtained in each case are presented in Fig 8.

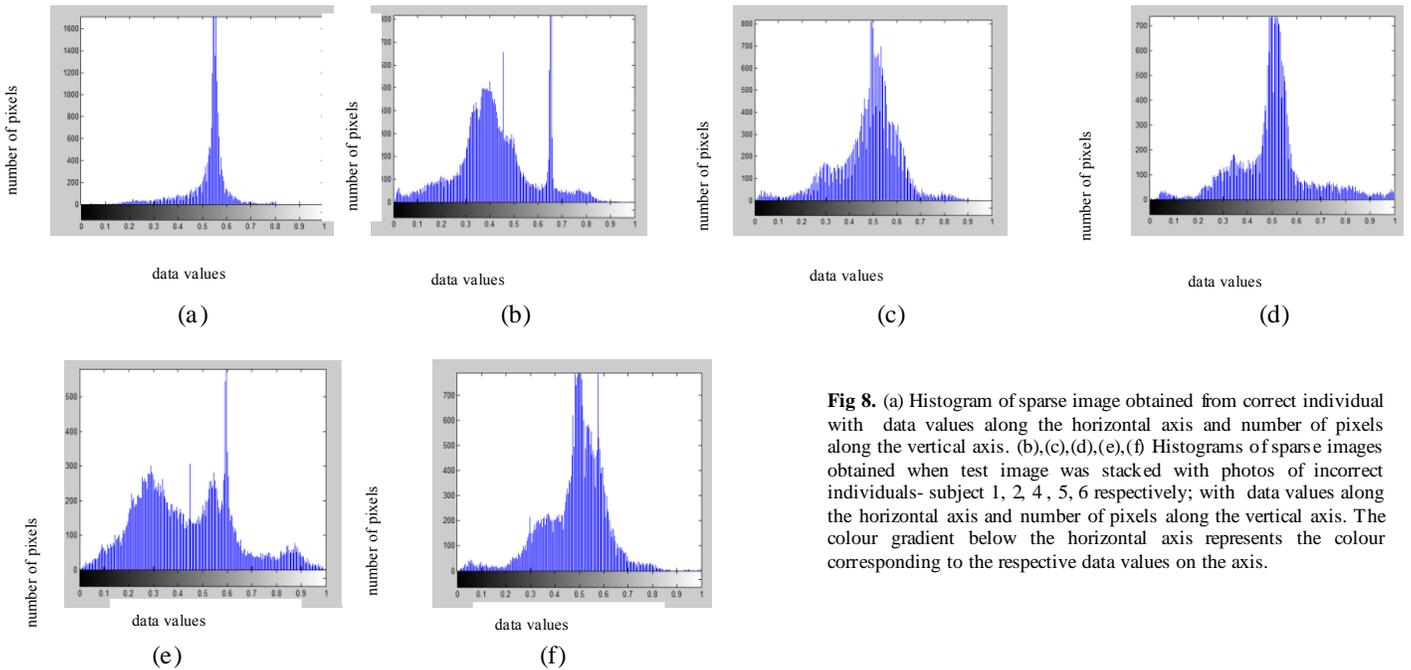

**Fig 8.** (a) Histogram of sparse image obtained from correct individual with data values along the horizontal axis and number of pixels along the vertical axis. (b),(c),(d),(e),(f) Histograms of sparse images obtained when test image was stacked with photos of incorrect individuals- subject 1, 2, 4, 5, 6 respectively; with data values along the horizontal axis and number of pixels along the vertical axis. The colour gradient below the horizontal axis represents the colour corresponding to the respective data values on the axis.

The histogram obtained from the sparse image of the correct individual (the person whose image is used as the test image) can be seen in Fig 8(a). It is observed from the peak in the histogram plot that maximum pixels have the data value between 0.5 and 0.6. The remaining colors are not much present in the picture, as it is clear from the histogram. These pixels between 0.5 and 0.6 mainly contribute to the occluded part that gets separated from the parent image to generate the lower-rank version. Subsequently, these pixels containing the occlusions depict the intra-class differences of the test image with the other 11 images (with which the test image was stacked). The histograms obtained from sparse images when applied to incorrect individuals are presented in Fig 8(b), (c), (d), (e), (f). As can be seen in these plots, many more prominent colors contribute to the images. For example, in Fig. 8(b), it is observed that majority pixels correspond to data values between 0.3 to 0.5 and again between 0.6 and 0.7. This happens because apart from the occlusions of the test image, the sparse components of person 1 are also taken up in the image like the test subject's face's edges, hair outline, etc. which overlaps with the intrinsic sparsity of test image. These sparse images, apart from the intra-class differences, also contain the inter-class differences. Similarly, in the plot of Fig 8(f), which happens due to the stacking of the test image on images of person 6, large number of pixels are seen to have data values between 0.3 and 0.7. Thus, it is observed that in these images, there is no single peak in the histogram plot, and different colors are seen to be distributed across the pixels of the images. Thus, by studying the histogram-plots of sparse images arising from test images when stacked onto different individuals' photos, several parameters can be coined to detect the correct individual to whom the test image belongs. These histogram plots can also be fed to machine learning algorithms, and training can be done on these test images so that the system can identify the image-plot having the least number of variant colors or data values that mainly contribute to the image. In other words, the histogram-plot can be identified which has the steepest peak amongst all, and that plot will be the one corresponding to the correct individual.

## CONCLUSION

Through both quantitative and qualitative study of performances of the two most promising LRMA algorithms-RPCA and WSNM, it was clear that WSNM surpasses RPCA in its performance in the removal of facial occlusions, thus giving recovered low-rank images of higher PSNR and SSIM. WSNM also shows consistency in its results and the ranks where best results occur in different occlusions, and different persons are not as random like those in RPCA. Testing these algorithms on our database, the same observations were made, making these conclusions more evident. As low-rank images sometimes might fail to capture the details of a face accurately, we propose the use of image-histogram of the sparse images (obtained when a test image is stacked with images of different persons) for facial recognition which is now the order-of-the-day in all surveillance and security purposes.